# Transfer Learning for Scientific Data Chain Extraction in Small Chemical Corpus with BERT-CRF Model

Na Pang[1,2], Li Qian[1,2], Weimin Lyu[3], Jin-Dong Yang[4]

[1] National Science Library, Chinese Academy of Science, Beijing 100190, China
[2] Department of Library, Information and Archives Management, University of Chinese Academy of Science, Beijing 100190, China
pangna@mail.las.ac.cn
[3] City University of New York, New York , USA
[4] Center of Basic Molecular Science (CBMS), Department of Chemistry, Tsinghua University, Beijing, 100084, China

**Abstract.** Computational chemistry develops fast in recent years due to the rapid growth and breakthroughs in AI. Thanks for the progress in natural language processing, researchers can extract more fine-grained knowledge in publications to stimulate the development in computational chemistry. While the works and corpora in chemical entity extraction have been restricted in the biomedicine or life science field instead of the chemistry field, we build a new corpus in chemical bond field annotated for 7 types of entities: compound, solvent, method, bond, reaction, pKa and pKa value. This paper presents a novel BERT-CRF model to build scientific chemical data chains by extracting 7 chemical entities and relations from publications. And we propose a joint model to extract the entities and relations simultaneously. Experimental results on our Chemical Special Corpus demonstrate that we achieve state-of-art and competitive NER performance.

## 1   Introduction

Recently, AI has stimulated the application of chemistry in many fields, such as computational chemistry and synthetic chemistry. Several tasks have highlighted the significance of the AI's role in chemistry. Scientists utilized deep neural networks and Monte Carlo tree to plan chemical syntheses and discover more retrosynthetic routes in short time[1], proposed machine learning method to perform chemical reactions and analysis faster than they could be performed manually and predict the reactivity of possible reagent combinations[2] and borrowed word2vec of NLP to create unsupervised machines Atom2Vec to predict materials properties[3]. There is no doubt that AI is revolutionizing our understanding on chemistry. In chemistry, especially in computational chemistry, though the chemical bond energy (pKa) is essential, most values existing in scientific papers are extracted by experts manually and there exists no work to try to extract the pKa with the method of NLP.

In particular, we consider three challenges in the application of scientific data chain extraction: (1) The existing corpora may not satisfy the aim of our task because they focus on general chemicals or drugs; (2) The popular chemical NER systems use the machine learning methods or deep learning methods, but it requires abundant data to train; (3) Unlike the start-of-art method to extract triplets $\{E1, relation, E2\}$, the entities are not confined in triplets and some of them are irrelevant to our relation extraction and some of them don not have 1:1 relation, but 1:n or n:1 relation. This difference makes extracting scientific chemical data chains significantly a tough task.

The first challenge is caused by corpus accessibility. Currently most experiments to extract named entities and corpora are in the field of biomedicine or life science which focus on extracting the chemical drugs. And the corpora may not be accessible, such as, PubMed corpus and Sciborg corpora[18]. Considering the need of automatically extracting chemical bond energy to promote the development in computational chemistry, and solving challenges of semantic problems and numerous unknown words, we create a new corpus of papers of chemical bond field.

The second challenge is caused by the ability of start-of-art deep learning architecture. The deep learning methods usually requires big data to train in order to get a better model, however the existing corpus for data chain extraction is not only hard to obtain but also in small scale. What's worse, most corpus focuses on other fields instead of chemical field. Considering this situation, we try also to use transfer learning method to relief the challenge by pre-training on large out domain corpus before training on chemistry in-domain specific corpus.

The third challenge is caused by the aim of our project and the characteristic of our corpus. In our project, we not only extract the entities which have relations, but also extract the irrelevant entities to aid researchers to read and confirm the right relations extracted by our system. And the multiple entities in one relation is more complex than the traditional triplets. For this reason, we construct our own tagging scheme to extract more extensive entities and also present a novel BERT-CRF model to extract name entity and relations simultaneously to avoid possible loss during above two tasks.

Our contributions: (1) We constructed a specific ChemBE corpus; (2) We utilize transfer learning on pre-training to make sure that we could have a competitive result on our minimal dataset; (3) We build a novel BERT-CRF joint model to extract entities and relations simultaneously and build our chemical scientific data chain.

## 2 Related Works

**Entity extraction and relation extraction.** Named entity extraction is a main subtask of information extraction. The common NER methods are based

3on rules, dictionaries, machine learning and deep learning. There are numerous experiments conducted in many fields[4–6].Relation Extraction is also a crucial task of information extraction. There are 4 types of methods of extracting relationships: fully supervised learning methods[7, 8], distant supervised learning methods[9], tree based methods[10] and joint learning with entity and relation methods[11]. These 4 methods can be classified into 2 models: pipeline models and joint models. The previous three methods are pipeline models which treat entity extraction and relation extraction as two separate tasks, and the last one regards them as one task[11].

In this paper, we focus on the joint learning method to learn entities and relations simultaneously. The joint learning model usually has two methods: parameter sharing[12, 13], and tagging scheme[11]. Parameter sharing model mainly utilizes the sharing parameters of the bottom layers and do different tasks via the upper layers. Tagging scheme model uses new tagging method to convert two tasks into one task and thus one end-to-end model can solve two tasks in the meantime.

**Scientific data extraction.** Except the traditional entities, there exists a lot of new trials to explore the possibility of extracting the scientific data in the scientific papers to mine the latent potential of scientific papers, such as extracting measured information from text to form a numeric value paired with a unit of measurement with the method of rules[14], utilizing CRF to extract numerical attributes from discharge summary records and SVM to associate correct relation between attributes and values[15].

There are also some works concerning chemistry field[15]. The most tasks relate to the chemical entities are in the biomedicine domain[15], since researchers do not have rich annotated data to learn in the field of Chemistry. For example, in the field of biomedicine, Xie J et al. proposed a method of Bi-LSTM network to extract to extract e-cigarette components[16]. Until 2015, BioCreative put forward CHEMDNER task to specially learn chemical entities and chemical formula[17].

But still, there are several problems about the chemical entity extraction: (1) As for corpora[18], they are mainly in the field of biomedicine ; (2) As for the techniques, the researchers are concentrated in machine learning in chemistry field and deep learning is only applied to biomedical field in English chemical corpora. Researchers have to extract all types features, thus the generalization ability is not strong. And also, we need mass of data to train the model. Therefore, we need to establish our own specific chemical corpus and apply some techniques to our small corpus.

**Transfer learning.** Transfer learning could help have better results on small dataset. Upstream unsupervised pre-training can help use less source and time to do the downstream tasks. There are two methods to apply the pre-trained language representations to downstream tasks: feature-based approach (eg, ELMO[19]) and fine-tuning approach (eg, GPT[20], GPT2[21],BERT[22]). Feature-based approach includes pre-trained representations as additional features into embeddings. Fine-tuning approach fine-tunes the pre-trained param-eters in the specific downstream tasks. In our work, we use BERT in upstream to do pre-training and CRF in downstream to fine-tune with the task-specific data.

## 3 Methods

### 3.1 Problem Statement

Our main task is to automatically extract the chemical bond energy values in chemistry field publications, since the pKa value is crucial in computational chemistry and well-build pKa values can pave the way for deeper research on computational chemistry. More specifically, we need to *extract 7 types of entities* and also *extract bond energy data chains* which contains many relations among 7 types of entities: *compound, solvent, reaction, method, chemical bond, Bond Energy(pKa) and Bond Energy value(pKa value), see figure 1. These 7 entities will construct a complete chemical bond energy value chain: XX compound has A reaction in B solvent to study the C chemical bond with D method, which pKa is E value.* Figure 2 shows the architecture of our method.

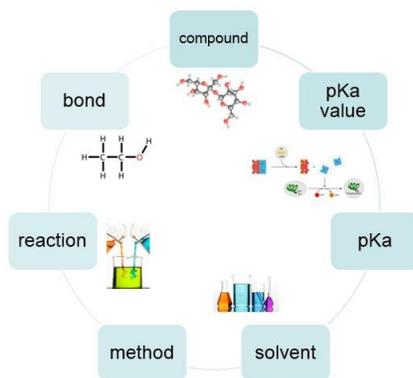

**Fig. 1.** 7 types of extracted entities

### 3.2 Construct Chemical Bond Knowledge Base and Corpus

We constructed a corpus of Chemistry papers annotated for NER task with the BIO encoding. We have 7 types of entities in our corpus: compound, solvent, method, reaction, bond, bond energy, bond energy value. *In our annotation work*, more than 10 experts of chemistry field validate the annotation results.

We invited chemistry experts from the Department of Chemistry of Tsinghua University and National Science Library of Chinese Academy of Science to construct our own chemical bond *knowledge base* and *corpus–ChemBE (Chemical Bond Energy) corpus. The knowledge base* includes dictionaries and rules, which are further used to recognize compounds and bonds later. The dictionaries include basic chemical formula and molecular formula of compounds, roots and affixes, radicals, substitutes, solvent, etc. The rules contain word indication rules, context indication rules and logical indication rules. *ChemBE corpus* is build up with 1900 full papers of chemical bond field following the process of gold standard corpus construction[23]. To ensure our corpus with high quality, multiple experts viewed the data independently and later inter-annotator agreement was needed to ensure quality. *ChemBE corpus* contains 7 entities: *compound, solvent, reaction, method, chemical bond, Bond Energy(pKa) and Bond Energy value(pKa value).* Table 1 shows the statistics of our chemistry corpus.

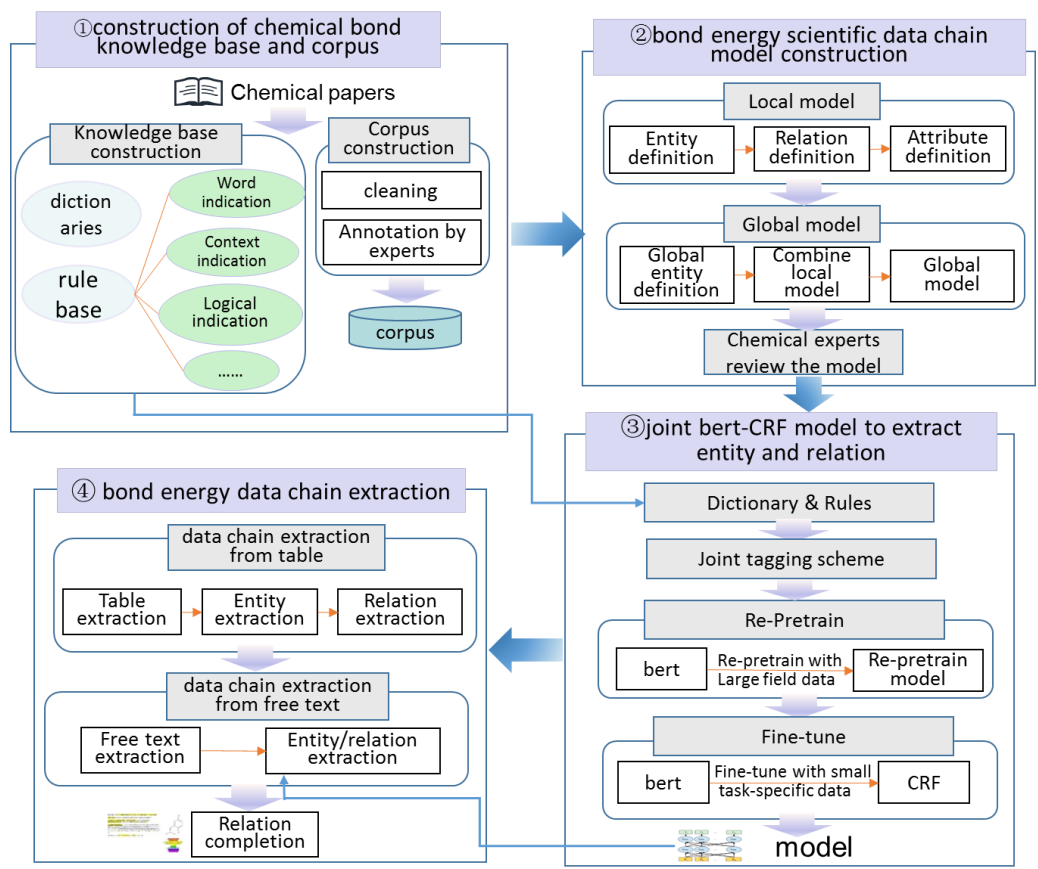

**Fig. 2.** process of our model

### 3.3   Bond Energy Scientific Data Chain Concept Model

Experts construct our bond energy scientific data chain model to assist our work. Experts build local model and global model to define the entities we need extract. There are 7 entities: compound, solvent, pKa, pKa value, bond, reaction and method. Among all the entities, we define 3 global entities(bond, reaction and method) and 4 local entities(compound, solvent pKa and pKa value). We only need to extract the relations between 4 local entities, since global entities can apply to the whole paper and we do not have to extract relations with global entities.

**Table 1.** Statistics of our chemistry corpus

| Entity Type | Total Num |
|---|---|
| Compound | 501722 |
| Solvent | 99013 |
| Reaction | 25841 |
| Method | 66124 |
| Bond | 19196 |
| Bond Energy(pKa) | 85312 |
| Bond Energy value(pKa value) | 14826 |

### 3.4 Joint BERT-CRF Model

In this part, we construct joint BERT-CRF Model to extract entity and relation simultaneously.

(1) Divide 7 entities into 2 categories and apply different methods to 2 types ( see Table 2).

**Table 2.** Annotated text corpora for training and assessment of chemical NER tools

| Type | Entity | Unknown words | Written form | Context | Method |
|---|---|---|---|---|---|
| Type1 | Compound, chemical bond | Much | Regular e.g., 1,8-Dihydroxy-4-naphthoic acid, O-H bond | High demand | Dictionaries and rules |
| Type2 | Solvent, method, reaction, pKa, pKa value | Little | Irregular | Low demand; polysemy (e.g., solution means solvent or answer) | BERT+CRF |

First, We use the established dictionaries and rules to replace compound and chemical bond entities with two marks: $CMP$ and $BOND$. Then, in the later deep learning process, we can avoid the unknown words trouble.

(2) Build our tagging scheme.

We build our own tagging scheme to extract both entities and relationships in the same time. In our tagging scheme, we only focus on only one relation between a pair of entities in our local models. Thus we define minimum relations between our local entities: compound-energy(CE) relation, solvent-energy(SE) relation and energy-energy value(EE) relation (see Figure 3). Among these relations, CE relation means "attribute", SE means "measure in" and EE relation means "the value of".

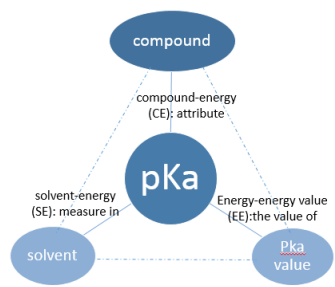

**Fig. 3.** 3 defined types of relations

And we define our tagging scheme like this (see Figure 4): <position information, entity information, relation information>. We give an annotation example (see Figure 5). The position information has 2 options: B and I, which means "begin" and "inter", respectively. The entity information has three options: com- pound, solvent and pKa value (the global entities and pKa entity not include, we only want to extract the relations of the other three local entities with pKa en- tity). The relation information has 4 options: CE (compound-pKa), SE(solvent- pKa), EE(pKa-pKa value) and NR(we only extract one relation among one pair of entities, thus we ignore other relations and all give them one tag <NR>, which means "no relation"). Other irrelevant words are tagged as <O>.

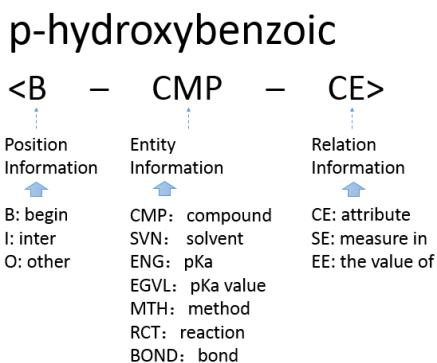

**Fig. 4.** tagging scheme

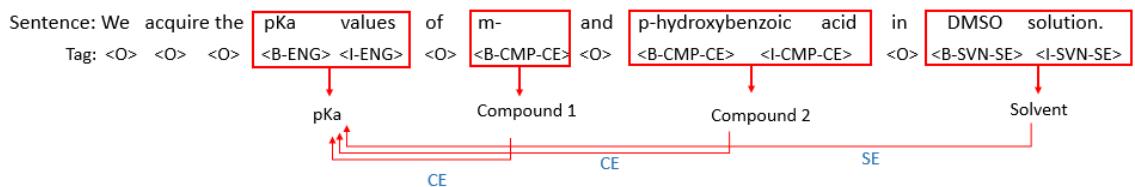

**Fig. 5.** annotation example

Thus, in our tagging scheme, when extract entities, <B-CMP-CE> and <B-CMP-NR> are equal, because we do not pay attention to the relations. In other words, we only pay attention to the first two parts in the tags. If an entity should be tagged as <B-CMP-CE>, we think <B-CMP-NR> extracts the correct entity, but the wrong relation.

(3) Re-pretrain BERT parameters with our large field data. We change the unused words in the vocabulary of BERT and re-pretrain the pre-trained parameters of BERT base with 700,000 abstracts in the field of chemistry.

(4) Fine-tune with small task-specific data. In the downstream NER task, we present a novel method by adding BERT softmax layer before adding a CRF layer to get better performance.

First, we use the BERT built-in softmax layer[22] to predict the labels. BERT defines two vectors in fine-tuning process: a start vector S and an end vector E. And during the fine-tuning process, we feed the final hidden representation $T_i \in R^H$ into classification layer and the we get a K dimensional vector, the possibility of the output vector belonging to category j is:

$$P_j(z) = \frac{e^{z^j}}{\sum_{i=0}^{k} e^{z^j}} \quad (1)$$

Then, we try to add CRF layer after BERT model to the downstream NER task. The CRF layer has a state transition matrix can use past and future tags to predict the current tag and scores possible tags to give a probability of the tag sequence. Given a sequence of input x={$x_1, x_2, ..., x_n$}, a sequence of predictions y={$y_1, y_2, ..., y_n$}, we define the score of the predictions as following:

$$S(x,y) = \sum_{i=0}^{n} A_{y_i, y_{i+1}} + \sum_{i=0}^{n} O_i, y_i \quad (2)$$

where A is a transition scores matrix, and O is the output matrix of BERT. We use our ChemBE corpus to train our BERT+CRF model (see Figure 6).

### 3.5 Extract bond energy data chain

(1) Extract data chain from table.

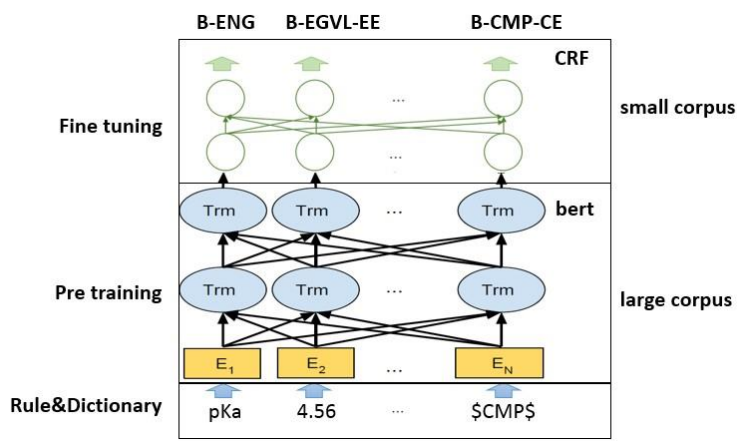

**Fig. 6.** joint BERT+CRF model

Tables always have some crucial entity and relation data. To some extent, extracting information from tables is not very tough, since tables have semi-structured data. We use dictionaries and rules to extract the entities and relations from tables.

(2) Extract data chain from free text.

We use our BERT+CRF model to predict the entities and relations in the free text.

(3) Complete the relations extracted from tables and free text.

Use entities and relation from the context and from the free text to complete our scientific data chain of pKa.

## 4 Experiments

**Entity extraction.** We conduct 4 different experiments to extraction chemical entities. We use two different downstream networks: softmax and CRF. We also use different parameters: parameters with only BERT pretraining and parameters with our re-pretraining with our chemical corpus. The results are shown in Table 3. We need to stress that as for compound and chemical bond entities, we use the dictionaroes and rules, not the deep learning method. We also make sta-tistical analysis of different entities of the most competitive model-BERT+CRF model (see Table 4).

**Table 3.** Entity extraction experiment results

| Settings | Task | P | R | F1 |
|---|---|---|---|---|
| BERT+softmax | Entity | 89.15% | 82.76% | 85.84% |
| BERT+softmax+re-pretrain | Entity | 89.69% | 83.11% | 86.27% |
| **BERT+CRF** | Entity | 91.56% | 87.27% | 89.56% |
| **BERT+CRF+re-pretrain** | Entity | **92.29%** | **87.48%** | **89.82%** |

**Table 4.** Experiment results of different types of entities of BERT+CRF model

| Entities | P | R | F1 |
|---|---|---|---|
| Compound | 86.67% | 84.78% | 85.71% |
| Bond | 87.65% | 84.52% | 86.06% |
| Method | 92.37% | 85.15% | 88.61% |
| Solvent | 91.52% | 85.03% | 86.16% |
| Reaction | 93.75% | 85.23% | 89.29% |
| pKa | 92.66% | 95.28% | 93.95% |
| pKa value | 91.67% | 87.30% | 89.43% |

As we can see in Table 3, our BERT+CRF model with re-pretrain parameters outperforms other models significantly. BERT+CRF model gains 3.72% improvement with no re-pretrained parameters and 3.66% improvement with re-pretrained parameters in F1 score, respectively. With re-pretrained parameters, BERT+softmax model gains an improvement of 0.43% and BERT+CRF model gains an improvement of 0.26%.

**Relation extraction.** We conduct 4 different experiments to extraction relations between different chemical entities. Like extracting chemical entities, we use two different downstream networks: softmax and CRF. And use different parameters: parameters with only BERT pretraining and parameters with our re-pretraining with our chemical corpus. The results are shown in Table 5.

**Table 5.** Relation extraction experiment results

| Settings | Task | P | R | F1 |
|---|---|---|---|---|
| BERT+softmax | Relation | 84.37% | 83.14% | 83.75% |
| BERT+softmax+re-pretrain | Relation | 85.79% | 84.30% | 85.04% |
| **BERT+CRF** | Relation | 87.82% | **86.16%** | 86.98% |
| **BERT+CRF+re-pretrain** | Relation | **88.46%** | 85.71% | **87.07%** |

Results of different types of relations of BERT+CRF model are shown in Table 6. In relation extraction part, BERT+CRF model also have a comparably competitive result than built-in softmax model. With no re-pretrained parameters, BERT+CRF model sees an improvement of 3.23% in F1 score. With re-pretrained parameters, BERT+CRF model improves F1 score from 85.04% to 87.07%. The precision and F1 score of BERT+CRF model with re-pretrained parameters are better than others. However, the recall of BERT+CRF model declines slightly with re-pretrained parameters, compared with no re-pretrained parameters.

**Table 6.** Experiment results of different types of relations of BERT+CRF model

| Relations | P | R | F1 |
|---|---|---|---|
| CE | 78.57% | 84.61% | 81.48% |
| SE | 89.77% | 85.86% | 87.78% |
| EE | 88.89% | 85.71% | 87.27% |

As we can see in Table 6, the CE relation is the toughest one among 3 relations. The reason behind this is that in our corpus, the compound is the entity of highest frequency. But the proportion of compound with CE relation is relatively small which requires high demand of contextual semantic information.

**Results presentation.** We display our entity extraction and relation extraction results as Figure 7. One color represents one type of entity, and arrows represent the relations between entities.

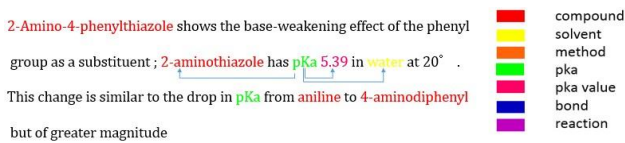

**Fig. 7.** Presentation of our results

## 5 Conclusions

We propose a joint BERT+CRF model to extract entities and relations simultaneously. The contribution of our work is threefold: (1) We investigate the performance of adding other task-specific network to downstream tasks of BERT. And the result shows that adding CRF to downstream NER tasks outperforms simple softmax. (2) We reference the multiple relation extraction of knowledge graph and propose our joint extraction model of entities and relations that only focuses on one relation between a pair of entities, but 1:n or n:1 entity pair in just one relation in one sentence. (3) We construct a model that could extract a chemical scientific data chain with multiple entities and relations.

## 6 Acknowledgements

The research work is supported by the Special foundation of Science and Technology Resources Survey (No.2018FY201202). We would like to thank the support by the Center of Basic molecular Science at Tsinghua University and National Science Library of Chinese Academy of Science. We thank Huizhou Liu, Li Qian, Jinpei Cheng, Jin–Dong Yang and Sanzhong Luo for the insightful suggestions and discussions.